# Biomimetic Use of Genetic Algorithms

Jean Louis Dessalles

TELECOM-Paris - Département Informatique, 46 rue Barrault - 75634 PARIS Cedex 13 - France - E-: dessalles@enst.fr

**Abstract:**
Genetic algorithms are considered as an original way to solve problems, probably because of their generality and of their "blind" nature. But GAs are also unusual since the features of many implementations (among all that could be thought of) are principally led by the biological metaphor, while efficiency measurements intervene only afterwards. We propose here to examine the relevance of these biomimetic aspects, by pointing out some fundamental similarities and divergences between GAs and the genome of living beings shaped by natural selection. One of the main differences comes from the fact that GAs rely principally on the so-called implicit parallelism, while giving to the mutation/selection mechanism the second role. Such differences could suggest new ways of employing GAs on complex problems, using complex codings and starting from nearly homogeneous populations.

## 1. GENES AND SCHEMATA AS THE UNITS OF SELECTION

A fundamental common feature between GAs and natural genetic systems comes from the fact that in both situations individuals are not what is selected.

In GAs, individuals are represented by their genome (most often a binary vector), and are evaluated so that only the best fitted ones have some chance to reproduce. The hope is that the GA will create individuals which will reach a maximum of the evaluation function. But this is not sufficient to make a GA: each pair of "parents" chosen among the best graded individuals enters a hybridation (crossover) in which genomes are mixed to give the genome of an offspring. This crossover is essential, as was shown by J. Holland. Without crossover we have random search, not GA. This requirement has many consequences for our purpose, as we will see after having compared it with its natural counterpart.

Figure 1 shows an example which was especially designed to underline the added value of crossover. Individuals are "marked" according to the sum of the bits of their genome. Crossover allows the selected 1s to gather in the same individuals.

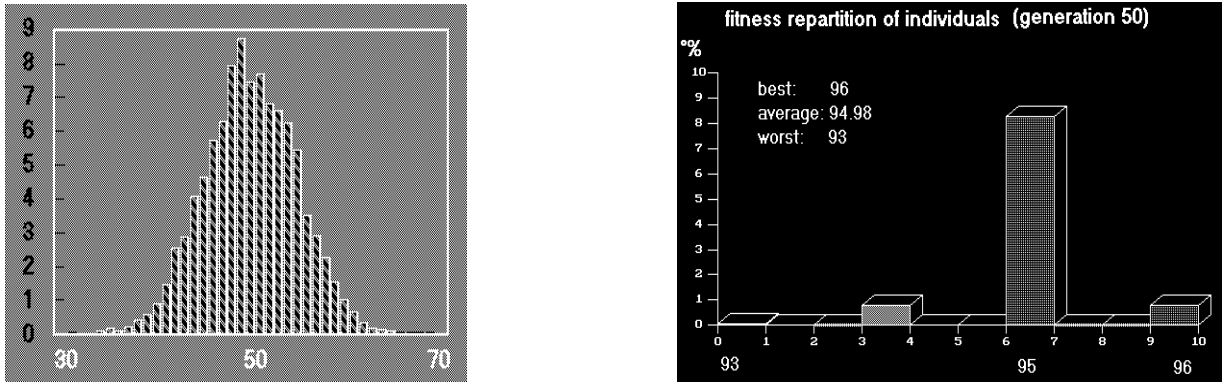

Figure 1. A genetic algorithm is asked to maximize the sum of 100 bits. The right histogram shows the result of 50 generations of 100 individuals: the whole population scores between 93 and 96. The left diagram shows the histogram resulting from 5000 random choices, which spreads between 34 and 66.

Crossover allows selection to operate not on individuals, but, as shown by J. Holland in 1968, on schemata (cf. [Goldberg 1989]). A schema is theoretically a partial specification of the genome, on which the observer is keeping an eye through generations. Practically, only schemata that are specified on a short length have some chance of survive after crossover. For the observer, only those schemata can pretend to be *selection units* . It would be indeed nonsense to say that the population as a whole is selected. And individuals are destroyed at each generation (if we except the so-called elitist models)! Only short schemata have a long enough life span so that, after some generations, some of them can be considered as selected by the observer.

J. Holland showed that at best $n^3$ different schemata were simultaneously present in a population of n individuals, and since schemata are selected when individuals are evaluated, he coined the term "implicit parallelism" ([Goldberg 1989]). But we must keep in mind that this counting of schemata is only valid for a random population.

This observation has to be compared with the ideas of Richard Dawkins, who pointed out that rigorously natural selection does not operate by selecting individuals, nor groups, but that it operates on genes (cf. [Dawkins 1978, 1982]). When considered from the point of view of natural selection mechanisms, individuals are mere temporary arrangements of genes. Only genes can survive through generations. Genomic hybridation is thus an essential phenomenon. Crossover is well known for sexual beings, but it has been also observed among unicellulars which reproduce normally through division: it is the so-called conjugation, during which two individuals mix their genome, and by the way no longer exist as individuals. There are reasons (from genetics and chemistry) to think that, contrary to what is usually said, crossover is as old as DNA, and that it is widespread among all living beings, although it has not been observed in each case. We could hardly speak of species in the absence of genomic encounters.

Since selection in GAs operates on schemata, one can define an average fitness value for each schema. But some schemata have also a meaning given by the designer, and it is often interesting to relate meaning and fitness, as a way to "explain" the solutions calculated by the algorithm. As we will see, this may not be possible in the case of biomimetic GAs.

## 2. BIOMIMETIC CONSTRAINTS ON THE GENETIC CODE

Using GAs is an art which lies for the major part in the coding of the problem. The designer has to define the semantics of the genome, so that a "mark" can be assigned to each individual. But the designer has to face several problems:

*non-separability*

- The "coding" of living beings is definitely complex. Natural selection operates on the phenotype of individuals, and the relation genome --> phenotype is very complicated (see for example [Petit 1976]): genes have multiple effects on the phenotype (pleiotropy) and phenotypic characteristics are the result of the combined action of numerous genes. Furthermore high order genes have been discovered (e.g. genes which code for the repressor of the activator of another repressor, cf. [Beardsley 1991]).

But why should the designer of a genetic algorithm choose a complex coding? An answer could be that a too simple coding often reveals a too simple problem, for which GAs may sometimes be superfluous. Since schemata are selection units, one may be led to consider sections of genome as independant units on which evaluation will operate. This was actually the case in the example of figure 1: each bit contributes to the global "mark" apart from the others. But this means that the problem, as far as it can be solved by such a GA, can be split into as many independent sub-problems as there are independently evaluated units.

However, interesting problems are often non-divisible (some problems may be separable, and GAs are precisely used to separate them, finding out independant schemata, but we are speaking here about complex problems that cannot be hoped to be separable). Schemata are indeed *selection units*, but they cannot be seen as *fitness units* if the problem is really complex, since the fitness of a given schema is necessarily highly dependent on the presence or absence of other schemata in the genome. In such a situation, the use of simple codings is unuseful· It may even be restricting, because it allows only a restricted exploration of the fitness landscape, as we will see. Complex codings, on the other hand, may be needed if we want the link genome ↑ evaluation to be complex, as is the case for example when one uses second order schemata, i.e.

schemata that modify the way other schemata are evaluated (e.g. a group of bits coding for the number of relevant bits in an other section of the genome).

### *continuity*

- one may think conversely that if a complex coding was chosen to handle a complex problem (with no hope that this problem may be separable), then this coding may not follow the *continuity principle*. This principle claims that genomes which are close for the Hamming distance (number of elementary mutations needed to transform one into the other) should have most of the time about the same evaluations. This is what makes offspring viable and allows some tolerance to mutations. Manderick [1991] measures this continuity with the autocorrelation of fitness when elementary mutations are performed. Unfortunately, it seems that the complex coding we are claiming from a biomimetic point of view has little chance to respect this continuity principle. If for instance a second order schema is even slightly mutated, then several other parts of genome will change meaning, and the global "mark" will be significantly changed.

So how can we explain that the natural system, which is so complex, is quite continuous? Part of the answer comes from the fact that genes are *extended*. Most of functional natural genes code for complex molecules, proteins. An elementary mutation will for instance change one aminoacid into another, which may eventually slightly change the chemical properties of the protein, e.g. diminishing its catalytic power or specificity. The effect will most probably go unnoticed, or will even be cancelled (more protein synthesis) at the level of individual fitness. The problem with many GA-codings is that they concentrate too much meaning on each bit. An "extended" coding should allow punctual mutations to have limited effects, and would thus allow complex codes, as required in biomimetic GAs, to be continuous.

### *preservation of semantics*

The balance between these two requirements: complexity and continuity, appears thus to be quite difficult to be hold. But the designer must also grant the *preservation of semantics* through genetic macro-operators like inversion, translocations or duplications. Such transformations do occur in nature (very seldom for some of them). They allow completely new genes to be created, and "cooperating" genes to become closer and to constitute longer functional genes (polygenes). In GAs, a complete "gene from locus" separation creates two orders: a semantic order, which never changes, and a topological order which may change through macro-operators and which bears neighbourhood relations that are used during crossover. But such a de-correlation between meaning and position makes offspring very unlikely to be functional after crossover. In the natural counterpart, major changes may go unnoticed if they occur in non-coding regions of DNA. And genes or polygenes may sometimes change location with no consequence on their own functioning or the functioning of their genetic environment. This is made possible by the presence of signals within DNA (e.g. starting position for transcription) that make groups of genes quite independent. To sum up, macro-operators are biomimetic, but complete gene/locus separation is not. Biomimetic GAs should instead make use of signals to make absolute position in the genome less relevant.

The choice of the genetic coding is crucial when we design a genetic algorithm to process a given problem, but, if we follow the biological metaphor, it appears also very difficult, since the coding has to meet the requirements of contradictory constraints: complexity (to allow better exploration of non-separable problems), continuity, preservation of semantics. Furthermore, such constraints upon the *structure* of codings may have consequences for the way we *use* GAs. They indicate that the classical way of using GAs is not very biomimetic, as we will see now.

## 3. COMPLEXITY IN THE CODING, NOT IN POPULATIONS

Classical genetic algorithms interbreed individuals that are completely different. In most applications, initial populations are indeed set up randomly. The GA is then considered to have completed its job when the population becomes nearly homogeneous. The end of convergence is sometimes handled differently because GAs become very slow: in a nearly homogeneous population, crossover looses its major role which is now played by mutation. New solutions are generated much more by the mutation/selection mechanism than by implicit parallelism.

However, nature shows us a quite different situation. Individuals that can be successfully crossed share at least 99% of their genome, and most differences are phenotypically neutral! This may contribute to explain why the tremendously complex coding from genotype to phenotype, with genes of high degree, remains compatible with the continuity principle even through crossover: offspring look like their parents because both parents are similar, and crossover creates few changes (quantitatively and qualitatively) when yielding new genomes.

This questions the biomimetism of GAs as they are classically defined. If the coding is complex enough (as is required by biomimetism), then a random initial population will consist of individuals that are unlikely to give viable offspring when crossed, because offspring will have very low fitness or no meaning at all. Of course the genome of individuals processed by classical GAs could be compared to the 1% of the significantly variable genes of a natural population (less than 1000 genes for mammals). But even in case of a sudden environmental change, very few of the alleles on these variable loci may become evolutionary meaningful, contrary to what is expected with GAs.

Genetic algorithms thus appear to be most biomimetic when they seem least interesting, i.e. at the end of convergence, when the population tends to become nearly homogeneous, and when the role of mutation dominates the rearrangement due to crossover (figure 2). At this stage, implicit parallelism no longer exists.

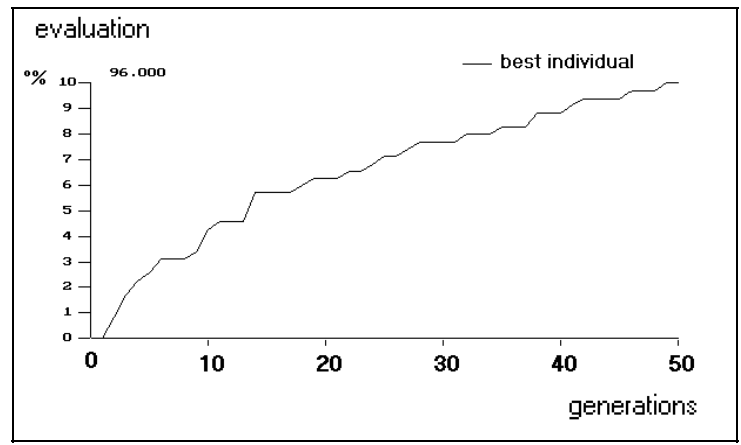

Figure 2. This schema shows that the GA of figure 1 converges asymptotically. The mere redistribution due to crossover does not allow to reach optimum (all bits 1). Admittedly we have a probability of only $[1 - (1 - 1/2^{100})^{100}] \cong 10^{-28}$ that one locus is 0 for all individuals in the initial population. But after the first generation, if 10% of the population reproduce, this probability reaches 0.1 . In the trial plotted above (elitist model), it seems that 4 loci remain 0 after 50 generations. If $p_m$ is the mutation probability, we have $[1 - (1 - p_m)^4] \cong 4.p_m$ chances that a given individual gains a 1 on one of these loci during the next generation. On can estimate by $1/10p_m$ the number of generations necessary to switch all four bits to 1. But increasing $p_m$ may be problematic, especially with a complex coding.

However what we want to suggest here is that the biomimetic use of genetic algorithms may be legitimate and could prove efficient in certain situations. Biomimetic GAs characteristics (as summarized in table 1) are legitimate because they bring the biological metaphor logically further. After all, GAs were not selected after a competition with all thinkable alternative algorithms. They were studied in the first place because of heuristic principles like "what performs well in nature should perform well in optimization". GAs, used biomimetically as proposed here, are of course less efficient. They lack the implicit parallelism put forward in most applications. However, they should appear as worth studying as well, if we can anticipate situations in which they could bring an added value, as may be the case with problems that cannot be formulated in terms accessible to usual methods.

Table 1
Main features of biomimetic genetic algorithms

| |
|---|
| ☐ complex coding, with high order genes, in order to allow better exploration of non-separable problems |
| ☐ extended genes to allow quasi-continuous mutational changes |
| ☐ use of signals (instead of complete gene-locus separation) to allow macro-operators |
| ☐ nearly homogeneous populations, where the mutation/selection mechanism takes the major role over implicit parallelism in the search for new solutions |

## 4. BIOMIMETIC GAs ON COMPLEX PROBLEMS

Good solutions for a complex problem are unlikely to be simple. Biomimetic GAs may be an interesting way to discover good complex solutions.

To examine this point we can consider the interesting attempt of Lavalou [1990] who tried to apply genetic algorithms to a complex enough problem. His genetic algorithm handled genomes coding for short music pieces which were evaluated by a connectionist network. He decided to design a complex coding with genetically encoded musical operators. For instance, musical themes were transformed recursively using short patterns, and not only the patterns, but also the number of recursive transformations were determined by genes.

The idea we want to suggest here in order to promote biomimetic GAs is that if the problem to be solved is complex (think of a very "tortured" and multi-dimensional fitness landscape), classical GAs will not perform better than random search! Crossover between necessarily very different (because initially randomly determined) individuals will give intermediary offspring which are very unlikely to obtain a significantly better fitness: if the fitness landscape is complex enough, each individual of the initial population is isolated in a local *fitness context* in which the fitness value of a given locus depends on the values of many other loci. Specific mutational changes may bring a higher fitness, while different (and often contradictory) changes would benefit another individual placed in a very different fitness context. In such a situation, crossover does not play its role of gathering independent good solutions together as expected with implicit parallelism, but it produces individuals, located in totally new fitness contexts, that can be considered as random in these contexts. The populations processed by classical GAs through generations correspond in this case to a few thousands of random trials, and if the problem is really complex, a satisfactory solution has little chance to be obtained this way.

Biomimetic GAs theoretically escape from this difficulty since they operate on nearly homogeneous populations. The initial random choice that locates the entire population near a given location in the fitness landscape will then be slowly, but constantly, improved by the combined effect of mutations and crossover. This is the mutation/selection mechanism. Then comes another feature of biomimetic GAs into play: the complexity of coding.

Since the link labelled *A* in figure 3 is very simple in classical GAs, the reachable phenotypes will have a quite limited complexity (e.g. algorithmic complexity as defined in [Chaitin 1987]) because of a lack of combinatorial power.

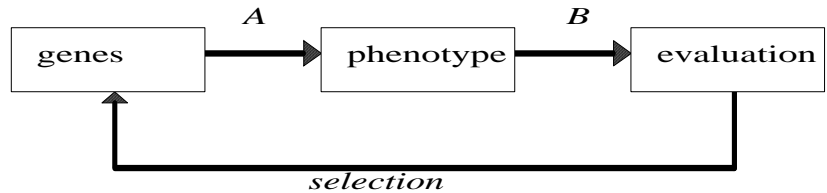

Figure 3. This schema illustrates the importance of the genotype/phenotype distinction for biomimetic GAs. This separation allows codings (in *A*) to create a "wide" range of phenotypes by complex combinations.

There are many ways to make the link *A* complex. For instance Werner & Dyer [1992] designed a system in which evolving creatures learn to communicate. In this application, the genome of each creature codes for a small connectionist system (its phenotype) that controls the creature's behavior. But we may think that combinatorial codes (i.e. with high order genes), as suggested for biomimetic GAs, may yield even richer genotype -- phenotype links.

If the coding leading from genes to phenotype is too simple (e.g. in the case of music, each note could be coded separately), then one can imagine that it is not an efficient way for the population to explore the landscape during its continuous walk (for instance a combinatorial coding of music pieces like the one chosen by Lavalou produces pieces of arbitrary length, what a trivial coding will never do). Simple codings do not have here the advantage of allowing a capture of parameters that are relevant to the problem (because the problem is not separable), and they may prove to be too limited in many complex applications.

## 5. POWER AND LIMITS OF BIOMIMETIC GAs

We may think of the genome of individuals processed by biomimetic GAs as a blueprint for a kind of machine (the phenotype of individuals). The mutation/selection mechanism has been invoked in very different contexts as a way to *create* complex systems (Darwin's natural selection theory, Burnet's clonal selection theory, synaptic selection theory of Edelman and Changeux, aspects of behavioral learning, and even cognition (see [Changeux & Dehaene 1989]), etc.). Can a system like biomimetic GAs create phenotypes that become increasingly complex? Can we imagine they could reach the same creative power as their natural counterparts?

Even if they should prove to be an interesting way to explore very complex problems, there are fundamental reasons why GAs as we know them are also limited in their creative power (this is particularly relevant to research on

Artificial Life, see [Jefferson et al. 1992]).

The semantics of GAs lies for the major part, and often totally, outside the genes. If we consider again Lavalou's program, a 5 bit sequence contributing, say, to a musical pattern has no chance to have any influence on the recursivity depth. In other words, one cannot imagine any combination of mutations, anywhere in the genome, that could change the semantics of these five bits this way.

In living beings, on the other hand, the semantics of genes is controlled by the genome itself, and it may be changed by the mutation/selection mechanism. If we discover that 5 codons have an influence on eye color, one could theoretically imagine a group of mutations somewhere else in the genome that makes these 5 codons control this time tooth length!

One can say that living beings are *endosemantic* entities, while GAs are *exosemantic*. The genome of living beings is the only known endosemantic system. If we define the "genetic envelope" of a GA as the set phenotypes that it can reach, then this enveloppe is predetermined (most of the time implicitly) by the designer of the genes --> phenotype coding. This may prevent us to expect too "miraculous" properties from GAs, even if they are biomimetic.

But GAs show quite impressive properties (generality, adaptation, learning capabilities when associated with classifiers,...). We tried to indicate here that, when used in a biomimetic way, they may be an interesting tool to explore very complex problems.

**acknowledgments:** I am grateful to Alain Grumbach, Irène Fournier and Olivier Hudry for having read an early version of this paper and for their fruitful remarks.

# References

Beardsley Tim (1991). La régulation des gènes. *Pour la Science* n°168, 54-64
Chaitin G. (1987). *Algorithmic Information Theory*. Cambridge Univ. Press
Changeux Jean-Pierre, Dehaene Stanislas (1989). Neuronal models of cognitive functions. *Cognition*, 33, 63-109
Dawkins Richard (1978). *Le gène égoïste (The Selfish Gene)*. Editions Menges
Dawkins Richard (1982). *The Extended Phenotype - The Gene as the Unit of Selection*. W.H. Freeman & Co, Oxford
Goldberg David E. (1989). *Genetic Algorithms in Search Optimization & Machine Learning*. Addison Wesley Publishing Company
Jefferson David, Collins Robert, et al. (1992). Evolution as a Theme in Artificial Life: The Genesys/Tracker System. In Langton C., Taylor C., Farmer J.D., *Artificial Life II*, Addison-Wesley
Lavalou Gilles (1990). *Application des algorithmes génétiques à la génération de séquences musicales*. Dossier long de fin d'études, ENST, Paris
Manderick B. (1991). *State of the art in theoretical approaches to genetic algorithms*. conf. 05 dec. 91 Alg. génétiques, ENS Ulm, Paris
Petit Claudine, Zuckerkandl Emile (1976). *Evolution: Génétique des populations, évolution moléculaire*. Hermann, Paris
Werner Gregory M., Dyer Michael G. (1992). Evolution of Communication in Artificial Organisms. In Langton Christopher, Taylor Charles, Farmer J.D., *Artificial Life II*, Addison-Wesley, 659-687